\documentclass{article} 
\usepackage{multirow}
\usepackage{nips13submit_e,times}
\usepackage{hyperref}
\usepackage{url}
\usepackage{natbib}
\usepackage{footnote}

\title{Distributional Models and Deep Learning
  Embeddings: Combining the Best of Both Worlds}

\author{
Irina Sergienya and Hinrich Sch\"{u}tze\\
Center for Information and Language Processing\\
University of Munich\\
Germany\\
\texttt{irina@cis.lmu.de}
}

\nipsfinalcopy 

\begin{document}

\maketitle

\begin{abstract}
There are two main approaches to the distributed
representation of words: low-dimensional deep learning
embeddings and high-dimensional distributional models, in
which each dimension corresponds to a context word. In this paper,
we combine these two approaches by learning embeddings based
on distributional-model vectors -- as opposed to one-hot
vectors as is standardly done in deep learning. We show that
the combined approach has better performance on a word relatedness judgment task.
\end{abstract}

\section{Introduction}
The standard approach to inducing deep learning embeddings
is to represent each word of the input vocabulary as a
one-hot vector (e.g., \citet{turian-ratinov-bengio:2010:ACL}, \citet{collobert:2011b}, \citet{DBLP:journals/corr/abs-1301-3781}). There is no usable information available in
this initial representation; in this sense, the standard way
of inducing embedding is a form of learning from scratch.

In this paper, we explore the question of whether it can be
advantageous to use a more informative initial
representation for inducing embeddings. Specifically, we
will test distributional-model representations for this
purpose -- where we define a \emph{distributional-model} or
\emph{distributional} representation of a target word $w$ as a
vector of dimensionality $|V|$;  the value on the
dimension corresponding to $v
\in V$ records a measure of the significance of the
dimension word $v$ for $w$. In the simplest case, this
measure of significance is a weighted cooccurrence count of
$w$ and $v$ (e.g., \citet{schutze92dimensions},
\citet{lund1996}, \citet{baroni10distributional}). 

Distributional representations have been
successfully used for a wide variety of tasks in natural
language processing, like synonym detection \citep{landauer97solution}, 
concept categorization \citep{Almuhareb04m.:attribute-based}, 
metaphorical sense identification
\citep{turney-EtAl:2011:EMNLP} and 
sentiment analysis 
\citep{Turney:2003:MPC:944012.944013}. For this reason, it is natural to ask whether they can
also improve the quality of embeddings.
We will realize this idea by presenting \emph{distributional
vectors} to the neural network that learns embeddings,
instead of presenting \emph{one-hot vectors}. We will refer
to these two modes of learning embeddings as
\emph{distributional initialization} and \emph{one-hot initialization}.

Our expectation is that distributional initialization will
be particularly helpful for rare words that occur in only a
few contexts. It is difficult for one-hot initialization to
learn good embeddings for rare words. In contrast,
distributional initialization should make the learning task
easier. For example, the distributional signatures of two
rare legal words will be similar since they occur in similar
contexts.  Thus, distributional
initialization will make it easier to learn similar
embeddings for them. In one-hot initialization, the
embeddings for the two words are initialized randomly and it
will be difficult for them to converge to close points in
the embedding space during learning if each only occurs in a
few contexts.

\section{Method}
\label{method}
As we just discussed, we would expect distributional initialization
to be beneficial mostly for rare words. Conversely, it is
likely that distributional initialization will actually
\emph{hurt} the quality of embeddings learned for frequent
words. The reason is that distributional initialization puts
constraints on the relationship between different
embeddings. This is a good thing for rare words as it
enforces similarity of the embeddings of similar rare  words. But it
can be harmful for frequent words that often have
idiosyncratic properties. For frequent words, it is better
to use one-hot initialization, which in principle does not
impose any constraints on the type of embedding that can be
learned.

Based on this motivation, we propose a hybrid initalization scheme:
all words with a frequency $f > \theta$ are
initialized with one-hot vectors, 
all words with a frequency $f \leq \theta$ are initialized with
distributional vectors, where the frequency threshold
$\theta$ is a parameter.

We test two versions of this hybrid initalization:
\emph{separate} and
\emph{mixed}.
Let $n$ be 
the dimensionality of the
distributional vectors, i.e., the number of words that we
use as dimension words, and $k$ the number of words with frequency
$f > \theta$. 
In the \emph{separate} scheme, the input
representation for a word is 
the concatentation of a $k$-dimensional vector and an
$n$-dimensional vector.
For a frequent word, the $k$-dimensional vector is a one-hot
vector and the $n$-dimensional vector is zero.
For a rare word, the $k$-dimensional vector is zero
and the $n$-dimensional vector is its distributional vector.
In the \emph{mixed} scheme, the input representation
for a word is an $n$-dimensional vector. It is a one-hot
vector for a frequent word and a distributional vector for a rare word.

In addition to the two separate and mixed hybrid schemes, we
also test non-hybrid distributional intialization.
In that case,
the input representation
for a word is an $n$-dimensional distributional vector for
frequent words as well as for
rare words.

\section{Experimental setup}
As training set for the word embeddings, we use parts
02 to 21 of the Wall Street Journal \citep{treebank}, a
corpus of about one million tokens and roughly 35,000 word types.

We used two word relatedness  data sets for evaluation:
MEN\footnote{http://clic.cimec.unitn.it/~elia.bruni/MEN} 
\citep{bruni-EtAl:2012:ACL2012}
and
WordSim353\footnote{http://www.cs.technion.ac.il/~gabr/resources/data/wordsim353/wordsim353.html} 
\citep{Finkelstein:2001:PSC:371920.372094}.
The two data sets contain pairs of words with
human-assigned similarity scores.
We only evaluate on the
2186 MEN pairs (of a total of 3000) and 303 WordSim353 pairs
(of a total of 353) that are covered by
our data set, i.e., both words occurred in WSJ.

We added to the continuous skip gram model \citep{DBLP:journals/corr/abs-1301-3781} of
word2vec\footnote{https://code.google.com/p/word2vec/} 
both one-hot and distributional
initialization.  We use hierarchical softmax,
set the size of the context window to 11, min-count to 0 (do
not discard words because of their low frequency), sample to 1e-3 (discard the words in the training set 
with probability $P(w) = 1-\sqrt{t/f(w)}$, where $f(w)$ is
the frequency of word $w$ and $t$ is a chosen threshold
\citep{NIPS2013_5021}) and embedding size
to 100.

We use a simple binary distributional model: Entry $1\leq
i\leq n$ in the distributional vector of $w$ is set to 1 iff
$v_i$ and $w$ cooccur at a distance of at most ten words
in the corpus and to 0 otherwise.
We test the methods for frequency thresholds
$\theta \in \{1, 2, 5, 10, 20, 50, 100, 1000\}$.

For each initializaton condition, we train 10 models. We measure Spearman correlation of the 
gold standard -- 
human-assigned similarity scores -- with 
cosine similarity scores between word
embeddings generated by our models, and report correlation averages for each initialization setup.

\section{Results and discussion}
\label{results}
Table~\ref{table1} gives averaged Spearman correlation coefficients between human and
  embedding-based similarity judgments
on MEN and WordSim.
Embeddings are produced by skip gram models with one-hot,
hybrid (mixed or separate) and distributional
initialization.
The threshold is varied for the two hybrid models
(column ``$\theta$''). Correlation coefficients are
given in the last two columns.

\begin{table}[t]
\caption{Averaged Spearman correlation coefficients between human and
  embedding-based similarity judgments
on MEN and WordSim.
Embeddings are produced by skip gram models with one-hot,
hybrid (mixed or separate) and distributional
initialization. 
The threshold $\theta$ is varied for the hybrid models.
The best correlation in each column is
bold. Correlations significantly better than one-hot
initialization are marked with a *.}
\label{table1}
\begin{center}
\begin{tabular}{l|r|rr}
\multicolumn{1}{c|}{initialization}
&\multicolumn{1}{|c|}{$\theta$}
&\multicolumn{1}{|c}{MEN}
&\multicolumn{1}{c}{WordSim}\\\hline\hline 
\multirow{1}{*}{one-hot} &&10.58 &17.31\\\hline
\multirow{8}{*}{mixed} 
 &1 	&*18.02			&*19.63\\
 &2 	&*\textbf{19.05}	&*\textbf{20.36}\\
 &5 	&*15.93			&16.86\\
 &10 	&11.43 			&16.39\\
 &20 	&10.95 			&11.47\\
 &50 	&8.22 			&3.06\\
 &100 	&11.52 			&6.62\\
 &1000 	&9.21 			&6.76\\\hline
\multirow{8}{*}{separate} 
 &1 	&*18.46 		&12.06\\
 &2 	&*15.90  		&13.58\\
 &5 	&*16.98			&*18.05\\
 &10 	&7.01 			&17.76\\
 &20 	&7.46 			&12.63\\
 &50 	&5.75 			&8.61\\
 &100 	&8.95 			&5.25\\
 &1000 	&7.12 			&7.19\\\hline
distributional &&5.69 &4.82
\end{tabular}
\end{center}
\end{table}

The main result is that the two hybrid initializations
outperform one-hot initialization
significantly\footnote{Student's $t$-test, two-tailed, $p<.05$}
 on both MEN and WordSim data sets for low values of $\theta$. 
This result is evidence that a hybrid initialization scheme
can be superior to one-hot initialization for words with
very few occurrences.

Hybrid initialization only does well for low values of
$\theta$. In general, as $\theta$ increases, performance
goes down. This trend reaches its endpoint for
distributional initialization, which can be interpreted as
hybrid initialization with $\theta = \infty$. The
correlations for distributional initialization are at the
low end of the range of performance numbers and are lower
than one-hot initialization.

The fact that distributional initialization
performs worse than hybrid initialization
confirms our initial hypothesis that frequent
and rare words should be treated differently.
Distributional initialization for all words -- including
frequent words -- 
imposes
harmful constraints on the embeddings of
frequent words; it is probably also harmful because it
links the embeddings of rare words to those of  frequent
words, which
makes it harder for the skip gram model to learn embeddings for
rare words. 

\section{Related work}
The problem of  word embedding initialization
was also addressed by \citet{le-EtAl:2010:EMNLP}.
They propose three initialization schemes. Two of
them, re-initialization and iterative re-initialization, use
vectors from prediction space to initialize the context
space during training. This approach is both more complex
and less efficient than ours. The third initialization
scheme, one vector initialization, initializes all word
embeddings with the same random vector: this helps to keep
rare words close to each other because vectors of rare words
are rarely 
updated. However, this approach is also less efficient than
ours since the initial embedding is much denser than in our
approach.

\section{Conclusion}
We have proposed to use a hybrid initialization for
learning embeddings, an initialization that combines the standardly used
one-hot initialization with a distributional initialization for rare words. Experimental
results on a word relatedness task provide tentative
evidence that hybrid initialization produces better
embeddings than one-hot initialization.

Our results are not
directly comparable with 
prior research on modeling word relateness judgments,
partly because the corpus we use has low coverage of the
words in the evaluation sets.
We also use a simple binary distributional vector
representation, which is likely to have a negative effect
on the performance of the embeddings.

In future work, we will test our models on larger corpora
and look at a wider range of
distributional models to produce results that are directly
comparable with other work on word relatedness.

{\bf Acknowledgments.}
We would like to thank Sebastian Ebert for his help with the code.

\bibliographystyle{elsarticle-harv}
\bibliography{iclr2014_sergienya}

\begin{thebibliography}{15}
\expandafter\ifx\csname natexlab\endcsname\relax\def\natexlab#1{#1}\fi
\expandafter\ifx\csname url\endcsname\relax
  \def\url#1{\texttt{#1}}\fi
\expandafter\ifx\csname urlprefix\endcsname\relax\def\urlprefix{URL }\fi

\bibitem[{Almuhareb and Poesio(2004)}]{Almuhareb04m.:attribute-based}
Almuhareb, A., Poesio, M., 2004. Attribute-based and value-based clustering:
  {A}n evaluation. In: EMNLP. pp. 158--165.

\bibitem[{Baroni and Lenci(2010)}]{baroni10distributional}
Baroni, M., Lenci, A., 2010. Distributional memory: A general framework for
  corpus-based semantics. Computational Linguistics 36~(4), 673--721.

\bibitem[{Bruni et~al.(2012)Bruni, Boleda, Baroni, and
  Tran}]{bruni-EtAl:2012:ACL2012}
Bruni, E., Boleda, G., Baroni, M., Tran, N.~K., 2012. Distributional semantics
  in technicolor. In: ACL. pp. 136--145.

\bibitem[{Collobert et~al.(2011)Collobert, Weston, Bottou, Karlen, Kavukcuoglu,
  and Kuksa}]{collobert:2011b}
Collobert, R., Weston, J., Bottou, L., Karlen, M., Kavukcuoglu, K., Kuksa, P.,
  2011. Natural language processing (almost) from scratch. Journal of Machine
  Learning Research 12, 2493--2537.

\bibitem[{Finkelstein et~al.(2001)Finkelstein, Gabrilovich, Matias, Rivlin,
  Solan, Wolfman, and Ruppin}]{Finkelstein:2001:PSC:371920.372094}
Finkelstein, L., Gabrilovich, E., Matias, Y., Rivlin, E., Solan, Z., Wolfman,
  G., Ruppin, E., 2001. Placing search in context: The concept revisited. In:
  WWW. pp. 406--414.

\bibitem[{Landauer and Dumais(1997)}]{landauer97solution}
Landauer, T.~K., Dumais, S.~T., 1997. Solution to {P}lato's problem: The latent
  semantic analysis theory of acquisition, induction and representation of
  knowledge. Psychological Review 104~(2), 211--240.

\bibitem[{Le et~al.(2010)Le, Allauzen, Wisniewski, and
  Yvon}]{le-EtAl:2010:EMNLP}
Le, H.~S., Allauzen, A., Wisniewski, G., Yvon, F., 2010. Training continuous
  space language models: Some practical issues. In: EMNLP. pp. 778--788.

\bibitem[{Lund and Burgess(1996)}]{lund1996}
Lund, K., Burgess, C., 1996. Producing high-dimensional semantic spaces from
  lexical co-occurrence. Behavior Research Methods, Instruments, \& Computers
  28~(2), 203--208.

\bibitem[{Marcus et~al.(1993)Marcus, Marcinkiewicz, and Santorini}]{treebank}
Marcus, M.~P., Marcinkiewicz, M.~A., Santorini, B., 1993. Building a large
  annotated corpus of {E}nglish: The {P}enn treebank. Computational Linguistics
  19~(2), 313--330.

\bibitem[{Mikolov et~al.(2013{\natexlab{a}})Mikolov, Chen, Corrado, and
  Dean}]{DBLP:journals/corr/abs-1301-3781}
Mikolov, T., Chen, K., Corrado, G., Dean, J., 2013{\natexlab{a}}. Efficient
  estimation of word representations in vector space. In: Workshop at ICLR.

\bibitem[{Mikolov et~al.(2013{\natexlab{b}})Mikolov, Sutskever, Chen, Corrado,
  and Dean}]{NIPS2013_5021}
Mikolov, T., Sutskever, I., Chen, K., Corrado, G.~S., Dean, J.,
  2013{\natexlab{b}}. Distributed representations of words and phrases and
  their compositionality. In: Advances in Neural Information Processing Systems
  26. pp. 3111--3119.

\bibitem[{Sch{\"u}tze(1992)}]{schutze92dimensions}
Sch{\"u}tze, H., 1992. Dimensions of meaning. In: ACM/IEEE Conference on
  Supercomputing. pp. 787--796.

\bibitem[{Turian et~al.(2010)Turian, Ratinov, and
  Bengio}]{turian-ratinov-bengio:2010:ACL}
Turian, J., Ratinov, L.-A., Bengio, Y., 2010. Word representations: A simple
  and general method for semi-supervised learning. In: ACL. pp. 384--394.

\bibitem[{Turney and Littman(2003)}]{Turney:2003:MPC:944012.944013}
Turney, P.~D., Littman, M.~L., 2003. Measuring praise and criticism:
  {I}nference of semantic orientation from association. ACM TOIS 21~(4),
  315--346.

\bibitem[{Turney et~al.(2011)Turney, Neuman, Assaf, and
  Cohen}]{turney-EtAl:2011:EMNLP}
Turney, P.~D., Neuman, Y., Assaf, D., Cohen, Y., 2011. Literal and metaphorical
  sense identification through concrete and abstract context. In: EMNLP. pp.
  680--690.

\end{thebibliography}

\end{document}